\documentclass[10pt]{article}

\usepackage[a4paper,textwidth=18cm,textheight=24cm,top=2.85cm, bottom=2.85cm, left=1.5cm, right=1.5cm]{geometry}

\usepackage{icdp2009}

\makeatletter
\long\def\@makecaption#1#2{%
\vskip\abovecaptionskip
\sbox\@tempboxa{#1. #2}%
\ifdim \wd\@tempboxa >\hsize
#1. #2\par
\else
\global \@minipagefalse
\hb@xt@\hsize{\box\@tempboxa\hfil}%
\fi
\vskip\belowcaptionskip}
\makeatother

\usepackage{lmodern}

\usepackage{graphicx}
\usepackage{times}
\usepackage{algorithm}
\usepackage{algorithmic}

\begin{document}
\noindent

\bibliographystyle{ieeetr}

\title{Self-Weighted Ensemble Method to Adjust the Influence of Individual Models based on Reliability}

\authorname{YeongHyeon Park \thanks{Correspondence author: yeonghyeon@sk.com}, JoonSung Lee, Wonseok Park}
\authoraddr{SK Planet Co., Ltd.}

\maketitle


\abstract
Image classification technology and performance based on Deep Learning have already achieved high standards. Nevertheless, many efforts have conducted to improve the stability of classification via ensembling. However, the existing ensemble method has a limitation in that it requires extra effort including time consumption to find the weight for each model output. In this paper, we propose a simple but improved ensemble method, naming with Self-Weighted Ensemble (SWE), that places the weight of each model via its verification reliability. The proposed ensemble method, SWE, reduces overall efforts for constructing a classification system with varied classifiers. The performance using SWE is 0.033\% higher than the conventional ensemble method. Also, the percent of performance superiority to the previous model is up to 73.333\% (ratio of 8:22).

\section{Introduction}
\label{sec:introduction}
Recently, human work exhaustion has eased by automated technology. The automated technology includes machine learning (ML) and deep learning (DL) these can conduct classification, detection, and some other task. 

The automated algorithm based on DL is adopted for deployment in varied areas such as medical, national defense, manufacturing, and some others. Taking a case of example for the manufacturing industry, it is possible to aggregate the status and result of products at high speed and high accuracy with lower effort compared to conventional technologies. Moreover, it can discover the insight from aggregated information and provide them as feedback to improve yield.

However, applying the DL simply does not always solve the problem easily and stably. As a simple example, the two risks are already known as over-fitting and under-fitting. In the over-fitting case, the trained DL model shows lower performance for the sample of unknown space or distortion.

In this paper, we propose the novel ensemble algorithm to ease the above limitation. Our novel ensemble method is composed in a form adjusting the influence via the reliability of each trained model involved in ensembling.

\section{Related work}
\label{sec:related_work}
The case of classification task, ensembling technique enables improve the accuracy and stability. When a single classifier is used, the error is directly propagated to the result, but mitigate the above problem is possible via ensembling multiple classifiers, cover the mistakes of one or a few classifiers. For this reason, cases of adopting the ensemble method in classification problems are increasing \cite{honsi2019ensemble}.

The ensemble technique can be divided into two cases as known as bagging and boosting \cite{breiman1996bagging} \cite{schapire1999boosting}. Bagging is like a voting system \cite{breiman1996bagging}. In the above method, each classifier votes via the output of its own, and the final result is determined based on the principle of majority voting. This method has a limitation that the low-performance model and high-performance model have the same influence on the final output. For solving the limitation, the system operator can assign a weight for each model but it needs extra effort to find optimal weights or make weight finding algorithm \cite{xia2011ensemble} \cite{alvear2019improving}.

Boosting is like a recursive method that trains $N$ models sequentially. The model ${M}_{n}$ is trained by assigning weights to data that the model ${M}_{n-1}$ cannot properly classify \cite{schapire1999boosting}. Assuming that the time consumption to train every single model is $T$, in the case of bagging can be trained in time $T$ via learning independently and simultaneously, but boosting takes as much time as $N \times T$ because it needs to be trained sequentially. 

\section{Proposed approach}
\label{sec:proposed_approach}
In this section, we present a novel simple ensemble method in consideration of the above limitations. We consider the bagging-based ensemble method based on the time-taking limitations of the boosting-based method.

\begin{figure}[ht]
    \begin{center}
		\includegraphics[width=0.65\linewidth]{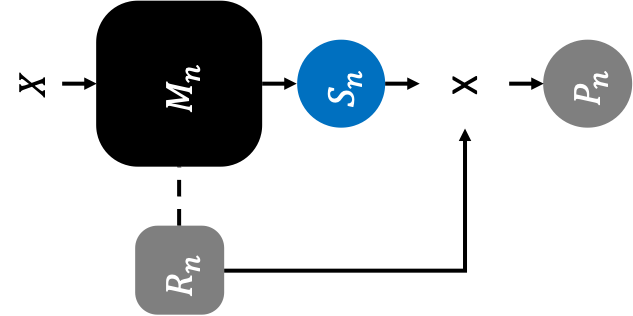}
	\end{center}
	\caption{The single module for constructing SWE.}
	\label{fig:module}
\end{figure}

\subsection{Single module for ensembling}
\label{subsec:single_module}
We name the proposed method as a self-weighted ensemble (SWE). The Table~\ref{tab:module} presents the symbol description for SWE module and Figure~\ref{fig:module} shows single SWE module. 

Our focus is to ease the cumbersome weight searching process in the bagging-based ensemble. The weight is determined by the performance index of the validation set after training. For example in the classification task, one of the indicators named F1-score can be used as a weight. 

\begin{table}[ht]
    \centering
    \caption{The symbol description of SWE.}
    \begin{tabular}{l|l}
        \hline
            \textbf{Symbol} & \textbf{Description} \\       
        \hline
            $X$ & Input data \\
            $Y$ & Output \\
            $M$ & Model to consist ensemble \\
            $R$ & reliability of single model \\
            $S$ & Classification score of single model \\
            $P$ & Partial output of single model \\
            $n$ & Model number $\{1, 2, ... N\}$ \\
        \hline
    \end{tabular}
    \label{tab:module}
\end{table}

\subsection{Self-weighted ensemble}
\label{subsec:swe}
\begin{figure}[ht]
    \begin{center}
		\includegraphics[width=0.90\linewidth]{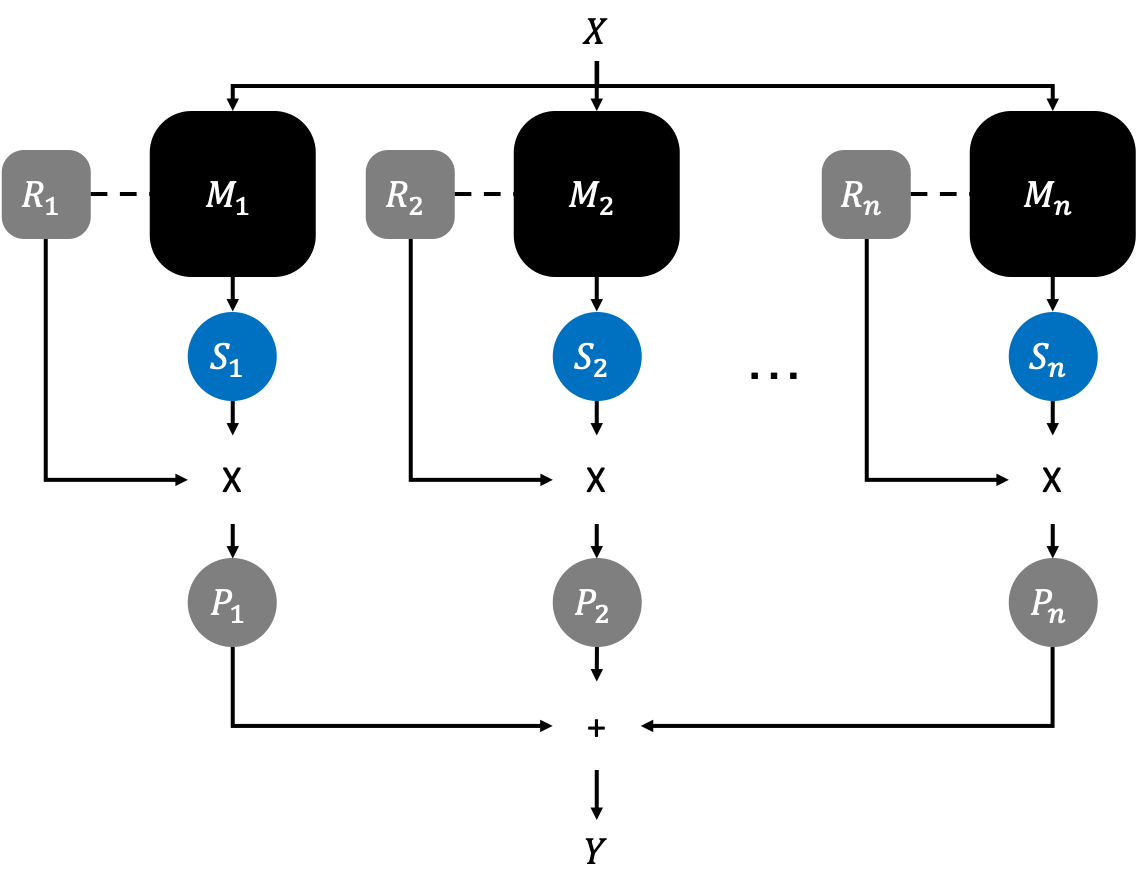}
	\end{center}
	\caption{The ensemble form of individual SWE modules.}
	\label{fig:ensemble}
\end{figure}

In the ensembling phase, several SWE modules take input data $X$ respectively. Then, the classification score is generated first via a trained model for each module, and a partial output is generated by multiplying with its reliability. The final output $Y$ is aggregated from the $N$ partial outputs as shown in Figure~\ref{fig:ensemble}.

The overall procedure for calculating the final output is also presented in Algorithm~\ref{algo:swe}. Through the algorithm, the summarized procedure that previously described can be confirmed.

\begin{algorithm}
	\caption{Procedure for generating the results in SWE}
	\begin{algorithmic} 
		\WHILE{End of training}
		\STATE Initialize and Train each model: $M_{n}$, $n=\{1, 2, 3, ... N\}$
		\STATE Measure the performance for each model: $M_{n}$ using validation set
		\STATE Save model $M_{n}$ and performance $R_{n}$
		\ENDWHILE
		\STATE Calculate socre for each score: ${S}_{n} = M_{n}(X)$
		\STATE Make partial output: ${P}_{n} = {S}_{n} * {R}_{n}$
		\STATE Summation for final output: $Y = \sum_{n=1}^{N}{P}_{n}$
	\end{algorithmic}
	\label{algo:swe}
\end{algorithm}

\section{Experiments}
\label{sec:experiments}
In this section, we deal with experiments to confirm the effectiveness of the SWE. The experiment is based on LeNet, a widely known form of convolutional neural network (CNN) \cite{lecun1998gradient}. Also, we evaluate whether the deep learning-based model has a consistent effect on the application of batch normalization (BN), a method that mitigates overfitting in single model training, and the depth difference of CNN \cite{loffe2015batch}. 

\subsection{Deep learning models}
\label{subsec:environment}

The LeNet consists of 3 layers by combining a convolutional layer (CL) and max pooling (MP) as a pair, and then 2 additional layers of fully connected Layer. 

We varied the above model via adding BN between CL and MP or constructing three layers as a pair of 2 CLs and 1 MP. For all models, only one layer was used for the fully-connected layer, unlike the basic LeNet. These three deep learning-based models are summarized in Table~\ref{tab:models}. 

\begin{table}[ht]
    \centering
    \caption{Three deep learning-based classification models constructed for the experiment. Each layer; convolution, max pool, fully connected, and batch normalization; is abbreviated as Conv, MP, FC, and BN respectively.}
    \begin{tabular}{l|ll}
        \hline
            \textbf{Model} & \textbf{Feature extractor} & \textbf{Classifier} \\       
        \hline
            LeNet-A & $\{Conv, MP\} \times 3$ & $\{FC\} \times 1$ \\
            LeNet-B & $\{Conv, BN, MP\} \times 3$ & $\{FC\} \times 1$ \\
            LeNet-C & $\{Conv, Conv, MP\} \times 3$ & $\{FC\} \times 1$ \\
        \hline
    \end{tabular}
    \label{tab:models}
\end{table}

\subsection{Classification performance}
\label{subsec:classification}
Experiments have conducted on the MNIST dataset using the three models presented in Table~\ref{tab:models}. Also, the experiment have repeated 30 times using each model using the Monte Carlo estimation \cite{kroese2014monte}. As a performance index, F1-score, the harmonic mean of precision and recall, is used. The measured average performance is summarized in Table~\ref{tab:f1score}.

\begin{table}[ht]
    \centering
    \caption{comparison of two ensemble methods using classification performance index F1-score.}
    \begin{tabular}{l|rr}
        \hline
            \textbf{Model} & \textbf{Bagging} & \textbf{SWE} \\       
        \hline
            LeNet-A & 0.98264 & \textbf{0.98274} \\
            LeNet-B & 0.98578 & \textbf{0.98601} \\
            LeNet-C & 0.98853 & \textbf{0.98886} \\
        \hline
    \end{tabular}
    \label{tab:f1score}
\end{table}

\subsection{Superiority check}
\label{subsec:superiority}
We also have counted the performance superiority of the bagging and SWE in each round of the repeated experiment process as shown in Table~\ref{tab:super}. In 30 repetitions, the number of situations in which the performance is superior among the two methods has counted. When the performance of the two models is the same, it has counted as a draw.

\begin{table}[ht]
    \centering
    \caption{The number of performance superiority in Monte Carlo estimation.}
    \begin{tabular}{l|rrr}
        \hline
            \textbf{Model} & \textbf{Bagging} & \textbf{Draw} & \textbf{SWE} \\       
        \hline
            LeNet-A & 9 & 3 & \textbf{18} \\
            LeNet-B & 6 & 1 & \textbf{23} \\
            LeNet-C & 8 & 0 & \textbf{22} \\
        \hline
    \end{tabular}
    \label{tab:super}
\end{table}

\subsection{Discussion}
\label{subsec:discussion}
Referring to the experimental results, the classification performance, F1-score, is improved from a minimum of 0.00010 to a maximum of 0.00033 depending on the model complexity. In an additional evaluation, the number of superiorities of SWE is larger than the bagging case. In the same context as the F1-score, the number of draws decreased in the case of model complexity increased, and the number of times the SWE winning increased.

\section{Conclusion}
\label{sec:conclusion}
In this paper, we propose a novel and simple ensemble method that every single model adjusts weight by its own reliability. Through our proposal, the effort of finding weights to apply for individual models is reduced. We also confirm that when the average performance with SWE is improved compared to the existing method, and over half of repeated experiments show a performance superiority.

\section*{Acknowledgements}
A preliminary version of the paper was presented at the 33rd Workshop on Image Processing and Image Understanding.


\end{document}